\documentclass[manuscript]{acmart}
\usepackage{cleveref}
\AtBeginDocument{%
  }

\setcopyright{acmlicensed}
\copyrightyear{2025}
\acmYear{2025}
\begin{document}

\title{Prompt-Driven Agentic Video Editing System: Autonomous Comprehension of Long-Form, Story-Driven Media}

\author{Zihan Ding}
\affiliation{%
  \institution{University of British Columbia}
  \city{Vancouver, BC}
  \country{Canada}}
\email{alex219@student.ubc.ca}

\author{Xinyi Wang}
\affiliation{%
  \institution{University of Bristol}
  \city{Bristol}
  \country{United Kingdom}}

\author{Junlong Chen}
\affiliation{%
  \institution{Department of Engineering, University of Cambridge}
  \city{Cambridge}
  \country{United Kingdom}}

\author{Per Ola Kristensson}
\affiliation{%
  \institution{Department of Engineering, University of Cambridge}
  \city{Cambridge}
  \country{United Kingdom}}

\author{Junxiao Shen}
\affiliation{%
  \institution{School of Computer Science, University of Bristol}
  \city{Bristol}
  \country{United Kingdom}}
\affiliation{%
  \institution{Memories.AI}
  \city{San Francisco, CA}
  \country{United States}}
\renewcommand{\shortauthors}{Ding et al.}

\begin{abstract}
Creators struggle to edit long-form, narrative-rich videos not because of UI complexity, but due to the cognitive demands of searching, storyboarding, and sequencing hours of footage. Existing transcript or embedding based methods fall short for video-editing workflows, as models struggle to track characters, infer motivations, and connect dispersed events. We present a prompt-driven, modular editing system that helps creators restructure multi-hour content through free-form prompts rather than timelines. At its core is a semantic indexing pipeline that builds a global narrative via temporal segmentation, guided memory compression, and cross-granularity fusion, producing interpretable traces of plot, dialogue, emotion, and context. Users receive complete edits end to end without the need for human intervention. Evaluated on 400+ videos with expert ratings, QA, and preference studies, our system scales prompt-driven editing, preserves narrative coherence, and balances automation with creator control.

\end{abstract}



\begin{teaserfigure}
  \includegraphics[width=1\linewidth]{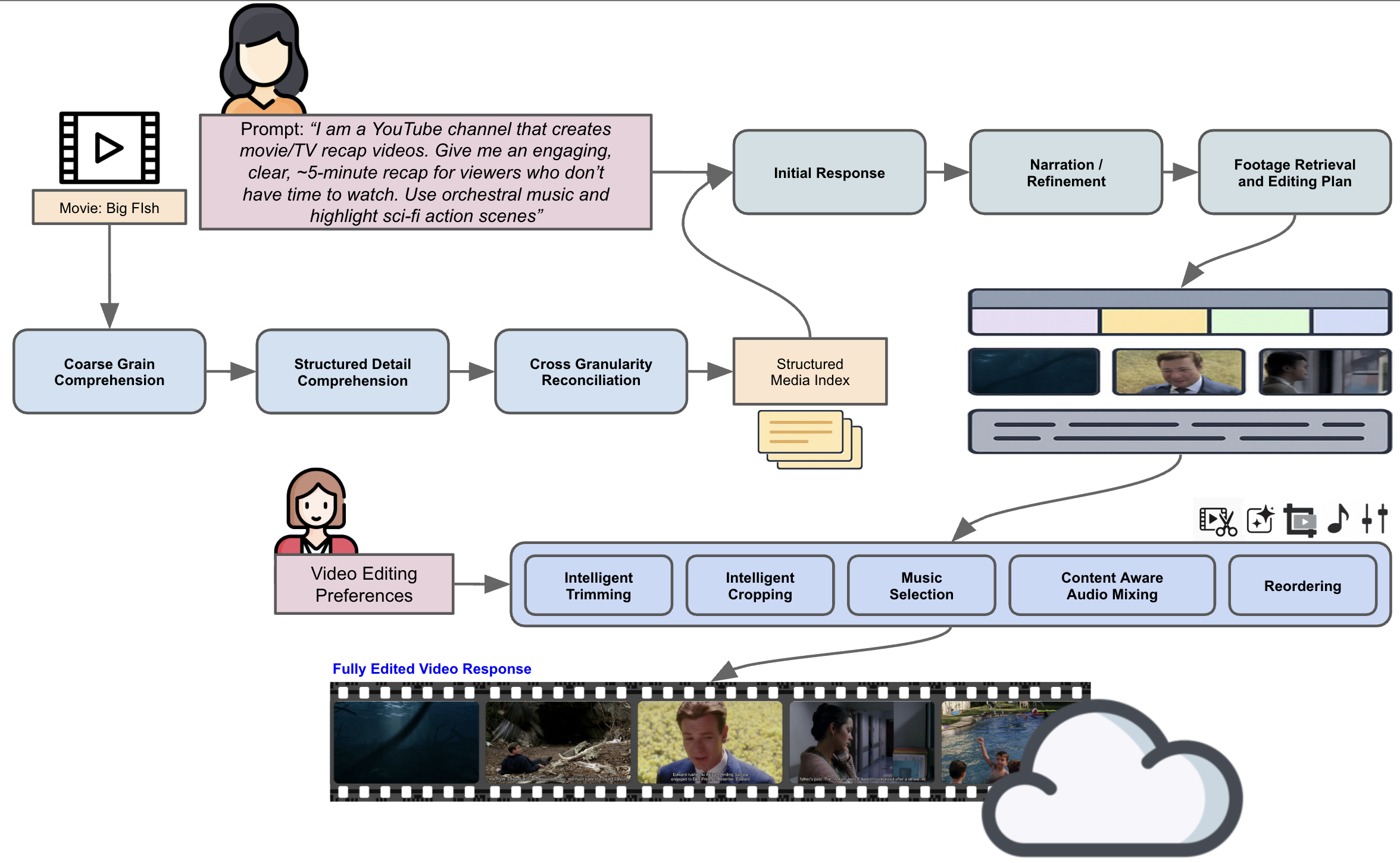}
  \caption{Overview of our AI editing pipeline for transforming long-form story-driven videos into concise, narrated cinematic recaps.}  \label{fig:teaserfigure}
\end{teaserfigure}

\maketitle

\section{Introduction}

Video editing is central to contemporary communication across professional studios, journalism, education, and independent creators. As projects become longer and more narratively complex, creators face a persistent bottleneck: shaping hours of footage into coherent stories requires searching, storyboarding, and fine-grained sequencing. Professional tools (for example, Adobe Premiere, DaVinci Resolve) offer timeline-level precision, and recent large multimodal models (LMMs) provide capabilities such as summarization and question answering. Yet these tools do not sufficiently automate the high-level, story-centric work of assembling highlights, reframing arcs, or retelling events across temporally dispersed scenes. Transcript or embedding based AI assisted workflows improves searchability, but does not yet achieve human level of comprehension, shifting cognitive load back to the user.

Recent models (for example, Gemini 2.5 Pro, GPT-4o) improve long-context reasoning, but still struggle with narrative abstraction and memory over multi-hour media. In our evaluations, models produce surface-coherent synopses yet lack timestamp reliability, character continuity, and causal grounding—limitations that make downstream editing and retrieval tasks fragile. At the same time, natural-language prompting has emerged as an expressive control surface for creators, enabling intent to be conveyed without timeline manipulation. This motivates a human-centered question: how might we combine free-form prompts with automated, reusable narrative understanding so that creators can direct long-form edits with low friction?

We present a prompt-driven, agentic video editing system that autonomously comprehends and restructures multi-hour, story-driven content while remaining interpretable. The system constructs a persistent semantic index by (1) hierarchical temporal segmentation, (2) guided scratchpadding with context compression, and (3) cross-granularity fusion between abstract summaries and timestamped scene traces. The index encodes plot points, character arcs, dialogue, affective cues, and visual context, all referenced by timestamps.

Interaction is entirely through natural language. Users can issue broad goals (for example, “Summarize this lecture as a 3-minute explainer”) or fine-grained constraints (“Exclude scenes with Character A, use somber background music, emphasize emotional transitions”), and the system responds with a coherent plan, interpretable intermediates (storyboards, narration scripts, edit plans), and a polished video. Modular agents for planning, retrieval, and rendering operate over the index, enabling parallel exploration of versions and styles, and supporting scalable, observable, and configurable workflows.

We stress-test the system on more than 400 long-form videos spanning narrative films, conference keynotes, interviews, and sports broadcasts. Expert ratings, structured question answering, and human preference studies assess semantic fidelity, narrative coherence, and usability. Our contributions are:
\begin{itemize}
    \item A modular, prompt-driven framework for autonomous comprehension and retelling of long-form, narrative-rich video.
    \item A robust semantic indexing pipeline that yields a reusable, time-aligned narrative representation (plot, characters, affect) suitable for editing and QA.
    \item An interpretable, agentic editing workflow that exposes intermediate artifacts for transparency, optional refinement, and parallel variant generation.
    \item A multi-pronged evaluation of our system in system reliability (Over 20x higher chance of receiving higher quality ratings compared to baseline models [Study 1], highest scores for correctness, timestamp accuracy, and detail compared to 3 baseline models [Study 2], and similar level of genre alignment and watchability compared to human-edited videos [Study 3]) and user experience (significantly more successful in long-form video recaps compared with two other professional video editing tools [Study 4]).
\end{itemize}

\section{Related Work}

\subsection{Intelligent and AI-Assisted Video Editing Interfaces}
Commercial editors increasingly surface AI within familiar workflows. Descript enables transcript-centric edits and overdub; Adobe Premiere Pro’s Text-Based Editing supports cut/assemble directly from transcripts; Runway and Pika provide prompt-driven generation and localized edits; and creation suites such as CapCut and OpusClip automate highlight extraction and social-formatting \cite{descript2025,premiere2025,runway2024gen3,pika2025,capcut2025,opusclip2025}. These tools reduce mechanical effort (silence removal, captions, reframing) but generally operate at the clip or asset level and do not maintain a persistent, editable representation of \emph{narrative structure} across hours of footage—limiting support for story-level operations such as arc tracking or causal reframing.

HCI research explores how AI and UI co-evolve during editing. Text-first or transcript-aligned systems lower the barrier for novices and improve control for experts: text-based editing for talking-head video demonstrates linguistically precise cuts \cite{fried2019text}; ExpressEdit combines \emph{language + sketch} for directable, region-specific edits \cite{tilekbay2024expressedit}; AVScript uses audio-visual scripts to make timeline manipulation accessible \cite{huh2023avscript}; and VideoMap reimagines browsing via a latent-space “map” to support ideation and discovery \cite{lin2024videomap}. Recent agentic interfaces bring LLMs into the loop: LAVE positions an LLM as a \emph{plan-and-execute} assistant that can brainstorm, search, storyboard, and trim while coexisting with timeline controls \cite{wang2024lave}; and VideoDiff presents \emph{alternative edits} for comparison and selection, supporting co-creative iteration \cite{huh2025videodiff}. This trajectory aligns with long-standing principles for mixed-initiative and adjustable autonomy in interactive systems \cite{horvitz1999mixed,parasuraman2000levels,amershi2019guidelines}. These systems foreground interaction quality and mixed-initiative control, yet they typically target short or well-structured content and do not maintain a persistent, time-grounded story model spanning multi-hour media. Our work complements this line by exposing an interpretable, timestamped \emph{narrative index} that both agents and users can operate on, enabling prompt-driven edits that remain coherent at feature-length scale.

\subsection{Language- and Diffusion-Driven Video Editing}
A parallel thread in vision leverages diffusion and instruction-following models for \emph{text-guided video editing}—altering style, lighting, or local content via prompts. Methods such as Video-P2P, TokenFlow, and Rerender-A-Video adapt image-editing controls to videos and improve temporal consistency via feature propagation or tracking; InstructVid2Vid and related pipelines enable zero-/few-shot edits driven by natural-language instructions; and StableVideo introduces consistency-aware diffusion for longer clips \cite{videop2p2023,tokenflow2024,rerender2023,qin2024instructvid2vid,chai2023stablevideo}. More recent techniques further improve editability and consistency across shots and scenes (e.g., Dreamix, FateZero, and one-shot tuning approaches) \cite{molad2023dreamix,qi2023fatezero,wu2023tuneavideo}. These systems are compelling for localized visual transformation, but they generally lack mechanisms for global narrative reasoning (e.g., maintaining character arcs, aligning edits to causality, orchestrating multi-scene beats). In our pipeline, such effects can be layered \emph{after} high-level planning: the edit plan is derived from a structured index (plot points, dialogue atoms, affect), then rendering agents may invoke generative effects while preserving story intent and timing.

\subsection{Long-Form Video Understanding and Narrative Modeling}
Long-form comprehension benchmarks foreground the difficulty of continuity, cross-segment memory, and causal reasoning. MovieNet, LVU, DramaQA, QVHighlights, and recent multi-task long-video suites emphasize narrative coherence and temporal grounding over hours of footage \cite{badamdorj2020movienet,lv2021lvu,kim2021dramaqa,qvhighlights2021,internvid2024}. Earlier datasets like MovieQA, TVQA, and LSMDC stress story-level QA and aligned descriptions \cite{tapaswi2016movieqa,lei2018tvqa,rohrbach2017lsmdc}, while MovieGraphs annotates social relations and motivations—useful for character-centric reasoning \cite{vicol2018moviegraphs}. Early pretraining approaches (VideoBERT, HERO) learned clip–text alignment but struggled with long-range abstraction \cite{sun2019videobert,li2020hero}; instruction-tuned audio-visual models (e.g., Video-LLaMA; M4) improved clip-level QA but remain bottlenecked by context windows and weak persistent memory \cite{zhang2023videollama,tsimpoukelli2023m4}. Newer foundation models (e.g., Gemini~2.5 variants) extend context and improve retrieval/localization, yet still exhibit gaps in narrative grounding—such as inconsistent character tracking, fragile temporal reasoning, and lack of reusable internal structure for downstream tools \cite{google2025gemini}. Our system addresses these gaps with guided scratch-padding and compression, hierarchical segmentation (macro-segments $\rightarrow$ scenes), and cross-granularity fusion to produce a reusable, time-aligned story representation that downstream agents—and users—can query and manipulate without re-ingesting raw video.

\subsection{Vector-Based Indexing and High-Speed Retrieval}
Vector search underpins scalable text–video retrieval and media RAG. Research systems (e.g., MemVid variants, uncertainty-adaptive and coarse-to-fine retrieval) and large corpora (e.g., InternVid) enable sub-second semantic lookup over massive archives \cite{chen2023memvid,tian2024coarsefine,internvid2024}. Large-scale pretraining datasets and retrieval models such as HowTo100M and CLIP4Clip/X-CLIP established strong baselines for text–video alignment at scale \cite{miech2019howto100m,luo2021clip4clip,ma2022xclip}. Proprietary platforms (e.g., Memories.ai) likewise embed segments to support natural-language search across long-form content \cite{memories2025}. While these approaches excel at isolated moment retrieval, they typically lack explicit temporal and causal structure: results may be relevant clips but not an ordered, narrative explanation. Our approach integrates vector retrieval where it shines (fast evidence lookup) and extends it with a structured, timestamped narrative index. This design supports tasks that require global story understanding—assembling highlights consistent with emotional trajectory, retelling from a character’s perspective, or reframing causality—while also producing interpretable intermediate artifacts (synopses, character graphs, edit plans) that can supervise or distill faster RAG systems.

\section{Intelligent Video Comprehension and Editing Pipeline}
We present a modular, cloud-native pipeline for long-form video comprehension and editing, centered around a hierarchical semantic indexing framework designed to enable robust interaction with large language models (LLMs). All media assets and derived comprehension artifacts are stored in Google Cloud Storage (GCS), enabling scalable, queryable access across tasks.

Our pipeline consists of three stages: (1) video comprehension and semantic indexing, (2) video-centric question answering, and (3) prompt-driven video response generation. All downstream modules operate over a structured, temporally grounded, and semantically enriched text index constructed via multi-pass comprehension of both coarse- and fine-grained temporal slices. We use Gemini 2.0 Flash as our primary vision-language model (VLM) due to its strong performance in video understanding tasks, though the pipeline itself is model-agnostic and compatible with any LLM with video comprehension capabilities.

\subsection{Interaction Flow}

Our system is designed to provide a seamless, natural language–driven user experience while maintaining robust automation and scalability. From raw footage to final cut, it only takes one prompt from the user. Users initiate a new project by uploading one or more video files to a dedicated project folder in Google Cloud Storage (GCS). This triggers a backend pipeline orchestrated by Temporal.io, which assigns workers to carry out comprehension and indexing tasks. 

Once the semantic index is constructed, users can issue freeform text prompts to perform tasks question answering and cinematic editing. These prompts may range from general (“Summarize the key points of this keynote”) to highly specific (“Create a trailer focusing on the emotional beats of the protagonist's arc”).

Intermediate artifacts generated by the pipeline—including the global synopsis, scene-level annotations, and clip selection plans—are exposed to users through HTTP APIs or UI endpoints, allowing for transparency. 

All editing outputs are uploaded back to GCS, with download URIs returned for viewing or publishing. The Temporal backend manages stateful orchestration, retries, and real-time debugging, enabling concurrent project execution and rapid iteration across multiple variants or prompts.

This architecture ensures that the system is not only autonomous and scalable but also interpretable and responsive to diverse user goals. It supports lightweight interaction without requiring users to scrub timelines or operate complex editing software, thus dramatically reducing the cognitive and operational burden of long-form editing workflows.

\subsection{Video Comprehension and Semantic Indexing}
To enable robust interaction with narrative-rich video, the system transforms multimodal input into a compact, interpretable textual representation. This occurs in three phases: rough comprehension, scene-level enrichment, and refinement.

\subsubsection{Temporal Segmentation and Pre-Processing }
Input videos are first decomposed into overlapping \textbf{15-minute segments} for global narrative construction, and finer-grained \textbf{5-minute scenes} for detailed semantic extraction. Segment overlaps ensure narrative continuity and mitigate failures arising from mid-dialogue or mid-action cuts. All segments are downsampled to 480p at 1 fps and compressed to minimize I/O overhead and reduce inference latency—aligning with Gemini 2.0 Flash's input constraints and performance characteristics.
\begin{figure}
    \centering
    \includegraphics[width=1\linewidth]{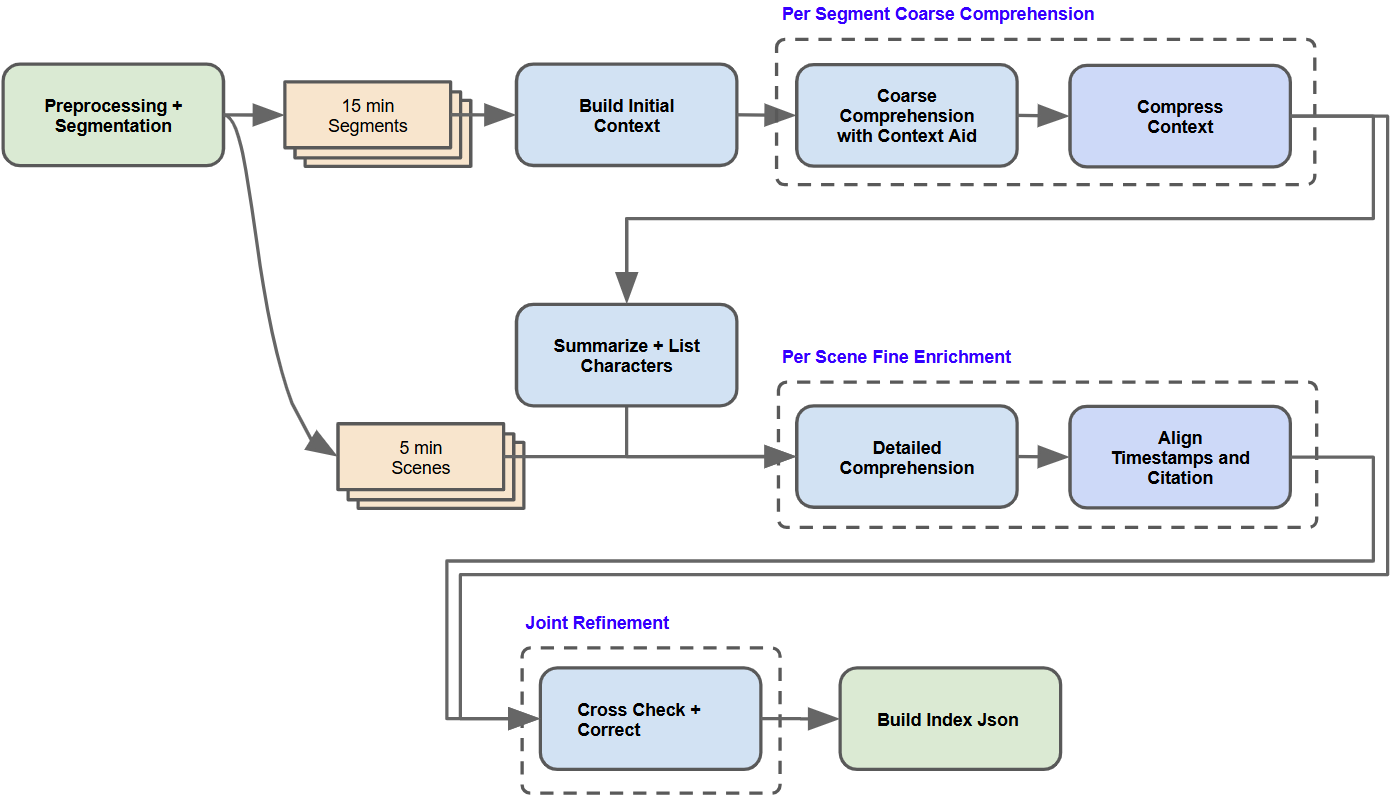}
    \caption{Workflow of Video Comprehension}
    \Description{}
    \label{fig:enter-label}
\end{figure}
\subsubsection{Coarse-Grained Comprehension}
The initial comprehension pass utilizes Gemini 2.0 Flash to process 15-minute segments, establishing high-level narrative scaffolding. For the initial segment, the LLM extracts the media format (e.g., cinematic, instructional), overall setting, story premise, key named characters, and their interpersonal dynamics. This structured summary acts as a persistent \textit{scratchpad} guiding subsequent segment interpretations.

To manage the strict token limitations of the LLM, naively accumulating context is impractical. We therefore implement \textbf{guided context compression}, wherein each segment's scratchpad is distilled into a compact, semantically meaningful summary using the LLM. These compressed summaries serve as efficient memory traces, enabling the LLM to effectively track evolving character arcs, plotlines, and relevant contextual facts without overwhelming the inference budget.

The output from this iterative comprehension pass includes a global plot synopsis draft and a character relationship graph, encoded as a textual adjacency map. Structurally constrained and consistently formatted prompts ensure deterministic outputs from Gemini 2.0 Flash, providing a stable foundation for downstream refinement.

\subsubsection{Fine-Grained Scene Comprehension}
Once a coarse global scaffold is in place, the pipeline proceeds to detailed semantic extraction at the 5-minute scene level. Without the global context, LLMs often struggle with speaker attribution, character continuity, and motivation inference, producing superficial or inconsistent interpretations. To address this, each scene is processed with the guidance of the draft synopsis, character relationship graph, and scene-level visual inputs.

Gemini 2.0 Flash is prompted to extract paraphrased dialogue, narrative-relevant speech acts, cinematographic descriptors (e.g., lighting, composition, setting), and affective signals (emotion, body language, tone). Semantic annotations are assigned timestamps at regular intervals (approximately every 20 seconds) or semantic boundaries such as camera cuts. The output is a high-resolution semantic trace serialized in structured JSON, facilitating precise temporal alignment and efficient querying in downstream tasks.

\subsubsection{Refinement Pass}
To reconcile and align local (scene-level) and global (synopsis-level) outputs, Gemini 2.0 Flash is prompted using both fine and coarse grain outputs to resolve missing character names, corrects inconsistencies or hallucinations, and enriches action descriptions with inferred causality. The final result is a comprehensive, self-auditing index containing a global synopsis, detailed scene breakdowns, and a refined character graph.

\subsection{Video-Centric Question Answering}
To support question answering, the system loads the index, formats it into a memory-efficient prompt, and pairs it with the user query. This avoids stochasticity common in direct multimodal prompting by anchoring reasoning in structured text. An auxiliary agent determines whether visual evidence is needed. If so, clips are retrieved using timestamp filters and semantic similarity matching. This supports precise answers to prompts such as \textit{“When did the protagonist express doubt?”} or \textit{“What were the consequences of that decision?”} 

 \subsection{Prompt-Driven Video Response Generation  }
This module enables cinematic video editing directly from natural language prompts by leveraging the structured narrative index produced during comprehension. Prompt-conditioned video editing should support high-level creative requests—such as summarizing specific arcs, extracting visual motifs, or retelling events from a character’s perspective—while maintaining narrative coherence and audiovisual polish. These requests are issued by users via freeform natural language prompts, with the system adapting its retrieval and rendering logic accordingly.

To meet this objective, our system distributes the editing workflow across a series of specialized agents. Each agent is responsible for a bounded subtask (e.g., narration planning, clip selection, rendering) and may invoke either an LLM (e.g., Gemini) or a domain-specific tool (e.g., FFmpeg, ElevenLabs, custom alignment heuristics). This modular design ensures low cognitive load per LLM invocation and enables precise orchestration via prompt engineering. We found that when LLMs are asked to both reason and produce structured output simultaneously, performance often degrades. Therefore, for complex tasks, we introduce intermediate freeform reasoning steps before issuing a second prompt to structure the result.

\begin{figure}
    \centering
    \includegraphics[width=1\linewidth]{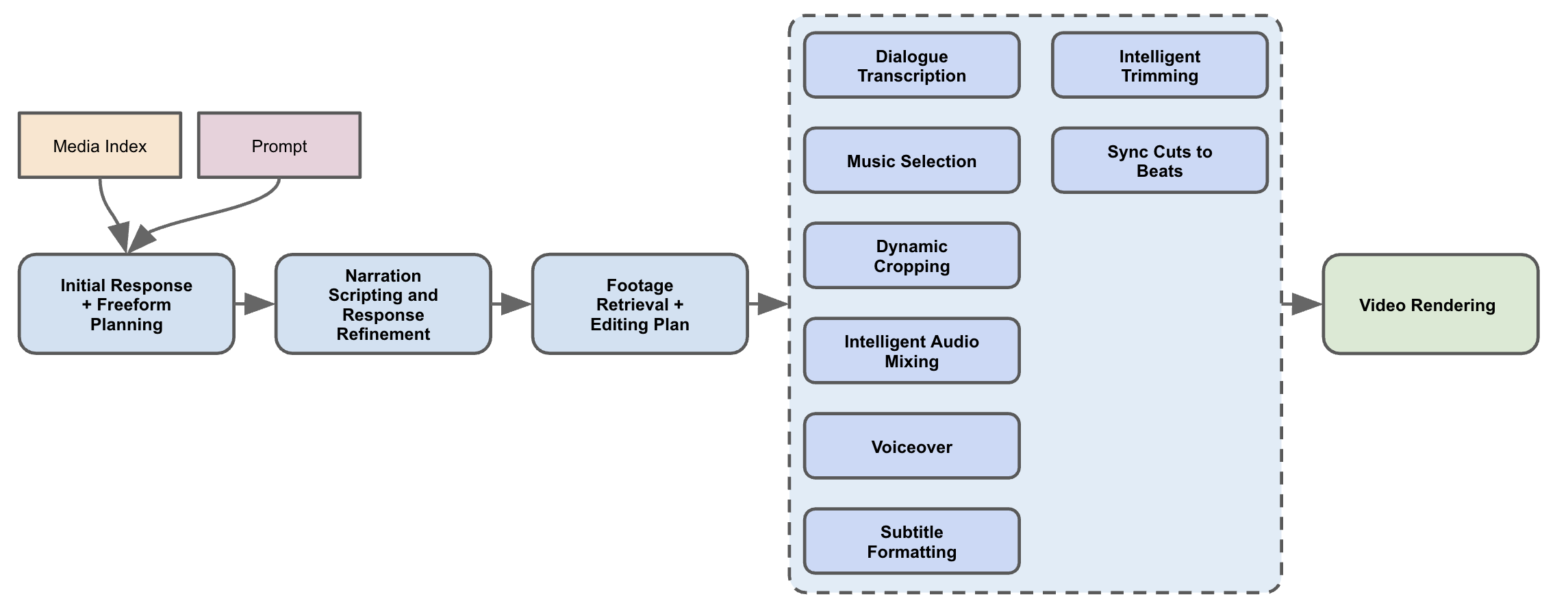}
    \caption{Workflow of Video Editing }
    \label{fig:placeholder}
\end{figure}
\subsubsection{Planning and Narration}
The planning and narration stage interprets the user’s prompt into a coherent storytelling framework that guides subsequent retrieval and rendering. This begins with a high-level planning agent that analyzes the tone, perspective, and scope implied by the user’s request. For instance, a prompt like “Retell the story from the antagonist’s point of view” demands not only content filtering but a deliberate shift in narrative framing. The planner prompts the LLM to consider these factors and generate a structured storyboard—a set of thematic or chronological segments outlining how the response should unfold.

When necessary, especially for prompts requiring complex causal reasoning or abstract interpretation, the planner first invokes a freeform LLM reasoning step to generate candidate ideas or decompositions. This is followed by a second prompt that formats the final storyboard into a structured plan usable by downstream agents.

Once the storyboard is generated, a narration agent converts each segment into a naturalistic voiceover script. This narration serves two roles: (1) it is user-facing, directly addressing or fulfilling the user's query in clear and engaging language; and (2) it provides a refined version of the storyboard, ensuring the video serves a central and consistent narrative purpose. These cues are invaluable for retrieval agents, which now have more grounded semantic anchors to search for corresponding visual scenes in the index.

\subsubsection{Clip Retrieval and Alignment}
The retrieval agent’s primary objective is to locate the most semantically relevant visual segments from the indexed video corpus and align them precisely with the narration plan. To ensure the highest possible alignment quality, the retrieval agent is prompted with the narration segment and instructed to concentrate its reasoning attention on the detailed, timestamped scene descriptions. The prompt is crafted to maximize token allocation on the indexed content, reducing cognitive overhead elsewhere and improving clip selection accuracy. This enables the LLM to reason globally across candidate scenes, favoring selections that most faithfully match the tone, pacing, and visual content implied by the narration.

The selected clips are serialized into a structured video editing plan—formatted as JSON—specifying time ranges, visual justifications, narrative function, and associated voiceover segments. This structure not only provides a deterministic scaffold for rendering but also supports auditability and post-hoc refinement.

Finally, each clip is dynamically assigned a rendering mode based on its narrative role and audiovisual clarity: muted video with narration overlay for exposition, raw audio for emotionally salient dialogue, or untrimmed inclusion for montage-style storytelling. This flexible rendering strategy ensures that each segment contributes effectively to the overall user-aligned narrative.
\begin{figure}
    \centering
    \includegraphics[width=0.6\linewidth]{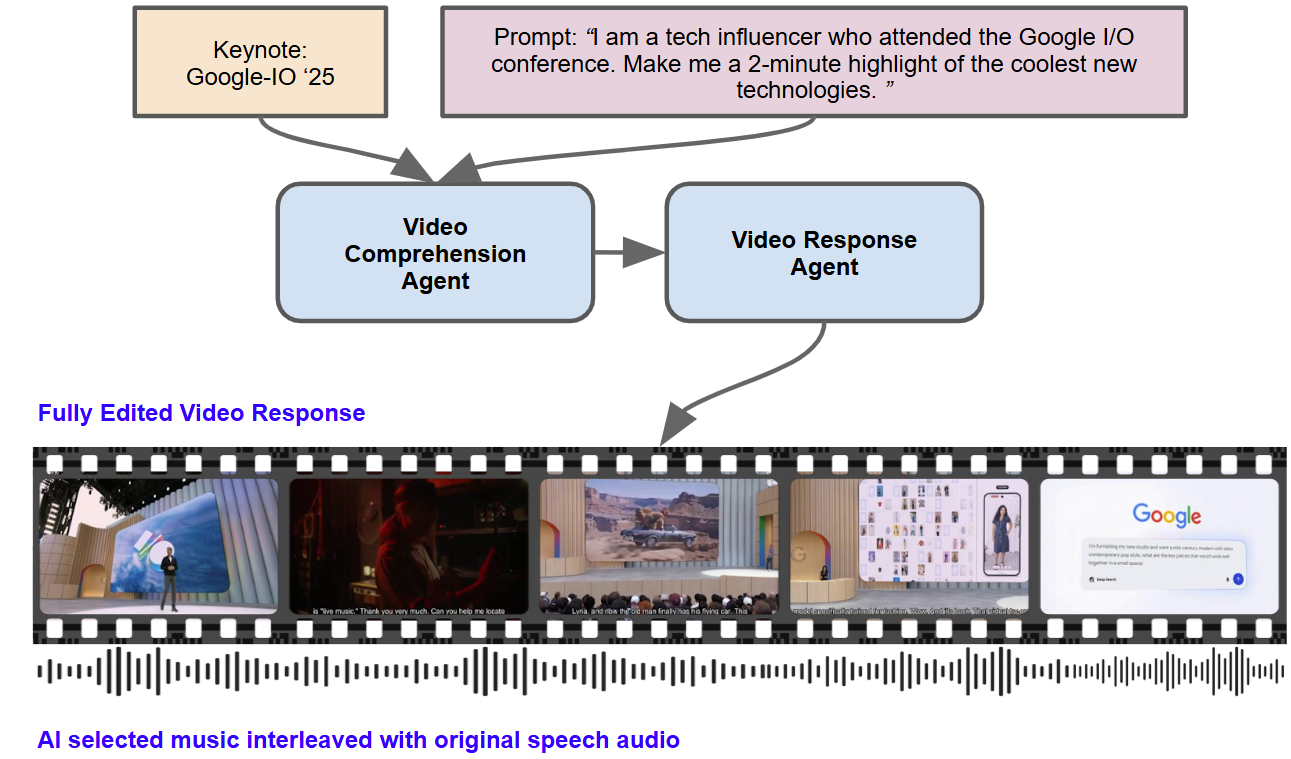}
    \caption{Example video editing request involving long-form content-rich footage}
    \Description{}
    \label{fig:enter-label}
\end{figure}

\subsubsection{Final Rendering}
The final rendering stage is orchestrated by a formatting agent that composes all retrieved video clips, narration audio, subtitles, and music—into a coherent cinematic output. We use FFmpeg and MoviePy as the core sequencing engines, with ElevenLabs TTS generating high-quality voiceover from the narration plan.

To support nuanced editing, several specialized agents handle less glamorous but crucial aspects of post-production. A beat-alignment agent uses Librosa for music beat detection and adjusts cut points to enhance rhythm and pacing—while aborting such alignment if it risks interrupting important narration. A dynamic cropping agent leverages pose estimation to track and frame key subjects, useful for mobile-friendly portrait modes or focused storytelling. Additionally, a micro-cut refinement agent uses ElevenLabs transcription to adjust cut boundaries at millisecond precision, ensuring spoken content is cleanly preserved and key expressions are not lost.

Subtitle overlays are aligned using heuristics based on narration timing and spoken content. Music selection is NLP-guided, with optional multilingual voiceovers, snap cuts, and aspect-ratio adaptation.

Together, these distributed, tool-augmented agents transform structured narrative representations into visually compelling, prompt-aligned video outputs that reflect both the user's intent and the source content’s internal logic. These capabilities enable users to explore creative directions, generate alternate versions, or produce stylistically tailored content with minimal manual effort.

\section{Evaluation}

\subsection{Study Overview}
We evaluated our system across three dimensions: (i) comprehension performance and reliability, (ii) quality of video editing outputs, and (iii) user experience and preferences. Our approach spanned four studies. Studies 1--3 involved quantitative expert ratings on 1--5 Likert scales with explicit rubrics, while Study 4 focused on qualitative user studies using structured questionnaires, open-ended interview prompts, and in-person observation. 

The studies were designed as follows. Study~1 tested the reliability and usability of media indexes for editing tasks. Study~2 assessed the richness and completeness of indexes in narrative media, including an ablation of our refinement stage. Study~3 measured the quality of edited videos, including whether outputs appeared ``AI-looking'' or fit genre expectations. Study~4 compared usability and perceived reliability against AI-enabled baseline editors. 

All participants provided informed consent, and procedures followed institutional ethics guidelines. Compensation was provided in the form of a small gift card. To ensure reproducibility, model versions, prompts, seeds, and configurations were logged, and all rater IDs and outputs were anonymized.

\subsection{Study 1: Reliability and Usability of Comprehension Indexes}

\paragraph{Goal.} 
We compared our system’s ability to generate structured media indexes with that of the Gemini family of models (2.0 Flash, 2.5 Flash, 2.5 Pro). The focus was on producing effective indexes for editing tasks, emphasizing reliability, structure, and detail. Movies and TV episodes were chosen because they are story-driven, long-context media that capture the complexity of general video.

\paragraph{Materials and Apparatus.}
The evaluation used 21 movies and 9 TV episodes. Indexes were produced by four systems: Gemini 2.0 Flash, Gemini 2.5 Flash, Gemini 2.5 Pro, and our system. Our system natively generates structured indexes as part of its comprehension stage. Gemini models required preprocessing: each movie or episode was split into 30-minute segments, since longer videos could not be processed reliably via the API. For each segment, Gemini was prompted to provide a detailed description of the content with as many scene-level timestamps as possible. A final consolidation prompt then merged the segment outputs into one unified index.

\paragraph{Participants.}
Eight experts between the ages of 20 and 35 participated. This convenience sample included social-media and video content natives, hobbyist creators, and technology professionals.

\paragraph{Procedure.}
For each media item, indexes were generated with all systems. Model identities were anonymized and the order of indexes was randomized. Raters were given the titles of the movies or episodes together with the evaluation rubric. If they were unfamiliar with a given title, they were explicitly permitted to search online (e.g., via Google) to gain sufficient context before scoring.

\paragraph{Criteria.}
All criteria were rated on a 1--5 scale, where 1 indicates an unacceptable or unusable index and 5 represents a perfect index. The criteria were divided into two categories:

\begin{itemize}
    \item \textbf{General quality.} Raters assessed whether the index was factually accurate, maintained a consistent formatting structure, and did not omit entire segments of the media.
    \item \textbf{Usability for editing.} Raters judged whether the index provided an appropriate amount of detail, contained a clear structure such that scenes could be looked up with corresponding timestamps, and whether those timestamps were accurate and consistent across the file.
\end{itemize}

\paragraph{Measures.}
Experts assigned scores for each criterion per index. All scores were logged and anonymized. The analysis procedure and results are described below.

\paragraph{Results.}

Our dependent rating variable was ordinal on a 5-point Likert scale and our dataset spans across 21 movies and 9 TV episodes. To account for the heterogeneity in comprehension difficulty across media files and the repeated measures from the eight experts, we analyzed the ratings using a Cumulative Link Mixed Model (CLMM) with random intercepts for both media file and rater. Fixed effects included \textsc{Model} (Gemini 2.0 Flash, Gemini 2.5 Flash, Gemini 2.5 Pro, and our proposed Agentic Pipeline) and \textsc{Category} (Quality and Usability).

The CLMM revealed a strong main effect of \textsc{Model} and a significant interaction effect of \textsc{Model}$\times$\textsc{Category}. Variability was moderate across media files ($SD=0.57$) but was negligible across raters ($SD<.0001$), suggesting that the 5-point Likert scale was used consistently across raters. Pairwise comparisons in \Cref{tab:posthoc} revealed that our proposed Agentic Pipeline (\textsc{Quality} $M=4.55, SD=.58$, \textsc{Usability} $M=4.45, SD=.62$) significantly outperformed the Gemini 2.0 Flash (\textsc{Quality} $M=3.19, SD=1.07$, \textsc{Usability} $M=1.10, SD=.29$) and Gemini 2.5 Flash (\textsc{Quality} $M=3.32, SD=.82$, \textsc{Usability} $M=1.13, SD=.34$) baselines in both \textsc{Quality} and \textsc{Usability} ($p<.001$ for all pairwise comparisons). Compared with the Gemini 2.5 Pro baseline (\textsc{Quality} $M=4.43, SD=.79$, \textsc{Usability} $M=1.29, SD=.51$), the Agentic Pipeline attained a significantly higher \textsc{Usability} rating ($p<.001$) but the difference in \textsc{Quality} rating between the two techniques was not significant ($p=.298$).

\begin{table}[h]
\centering
\caption{Pairwise comparisons of the proposed Agentic Pipeline against baselines with log-odds estimates, standard errors (SE), $z$-ratios, adjusted $p$ values (Holm correction), and odds ratios (OR). Significant differences are shown in \textbf{bold}.}
\label{tab:posthoc}
\begin{tabular}{llrrrrr}
\toprule
Category & Contrast & Estimate & SE & $z$ ratio & $p$ value & OR \\
\midrule
Quality   & Gemini 2.0 Flash -- Proposed & -3.265 & 0.203 & -16.063 & $\textbf{<.001}$ & 0.038 \\
Quality   & Gemini 2.5 Flash -- Proposed & -2.979 & 0.195 & -15.257 & $\textbf{<.001}$ & 0.051 \\
Quality   & Gemini 2.5 Pro -- Proposed   & -0.195 & 0.188 & -1.040  & 0.298  & 0.823 \\
Usability & Gemini 2.0 Flash -- Proposed & -9.227 & 0.357 & -25.832 & $\textbf{<.001}$ & $\approx$0.000 \\
Usability & Gemini 2.5 Flash -- Proposed & -8.958 & 0.343 & -26.091 & $\textbf{<.001}$ & $\approx$0.000 \\
Usability & Gemini 2.5 Pro -- Proposed   & -7.931 & 0.310 & -25.604 & $\textbf{<.001}$ & $\approx$0.000 \\
\bottomrule
\end{tabular}
\end{table}

We further computed pairwise odds ratios (ORs) to compare each baseline to the proposed system to support interpretation. The original model contrast was defined as <Baseline>-<Proposed>, so the inverted value $1/OR$ reflects <Proposed> relative to <Baseline>. For \textsc{Quality}, the odds of the Agentic Pipeline receiving a higher rating were approximately 26 times greater ($1/.038$) than Gemini 2.0 Flash and 20 times greater ($1/.051$) than Gemini 2.5 Flash, and about 1.2 times greater than Gemini 2.5 Pro ($1/.82$), though the latter difference was not statistically significant.
For \textsc{Usability}, the odds ratio approached infinity as the baseline ORs were near zero, which suggests that experts consistently rated the proposed Agentic Pipeline more usable than all baselines.

Together, these results suggest that our proposed Agentic Pipeline significantly outperforms all three baselines in structured usability, and significantly outperforms Gemini 2.0 Flash and Gemini 2.5 Flash but not Gemini 2.5 Pro in terms of narrative quality. 


\subsection{Study 2: Richness and Narrative Retention (with Ablation)}

\paragraph{Goal.}
This study evaluated whether indexes captured intricate plot details and more abstract narrative elements such as character intent, thematic emphasis, or authorial tone. We also examined the contribution of our system’s refinement stage by performing an ablation experiment. Movies were again chosen due to their story-driven nature and long duration.

\paragraph{Materials and Apparatus.}
Indexes were generated by our system (with and without refinement) and by Gemini 2.0 Flash, 2.5 Flash, and 2.5 Pro. To evaluate these indexes, we curated a set of 39 challenging questions across 8 movies. The questions were deliberately designed to test multiple aspects of comprehension: some asked for fine-grained factual details, others required reasoning about the chronological order of events, and others directly requested events tied to specific timestamps.

\paragraph{Participants.}
Twenty experts between the ages of 20 and 50 participated. This convenience sample included individuals with backgrounds in film, literature, and fine arts, many of whom had academic or professional interests in narrative media.

\paragraph{Procedure.}
For each (index, question) pair, Gemini 2.0 Flash was used as a uniform answering model. The model was prompted with both the index and the question, and explicitly instructed to answer using only the contents of the index while providing as much detail as possible. For our system, answers were generated using both the full and ablated indexes to assess the effect of refinement. Experts reviewed anonymized Q/A pairs alongside the ground truth answers before assigning scores.

\paragraph{Criteria.}
Scores were assigned on a 1--5 scale, where 1 indicates a completely incorrect or inadequate answer and 5 represents a perfectly correct and complete answer.

\begin{itemize}
    \item \textbf{Factual correctness.} Raters judged whether the answer contained the core facts matching the ground truth and whether any additional information was accurate rather than hallucinated.
    \item \textbf{Temporal accuracy.} If the question asked for a timestamp, raters checked whether it was correct. For questions about ordering, raters evaluated whether the sequence of events was correct.
    \item \textbf{Completeness.} Raters considered whether all parts of the question were answered, whether answers provided sufficient detail, and whether necessary context was included.
\end{itemize}

\paragraph{Measures.}
Experts scored all anonymized Q/A pairs according to the rubric. Ratings were logged and anonymized. The analysis procedure and results are described in the following section.

\paragraph{Results.}


To assess comprehension quality of the media files, we also constructed a set of question-answering (QA) keys for a set of eight diverse movies in the dataset. We prompt Gemini 2.5 Flash to answer these questions based on the structured data of each movie generated by baseline models (models 1 to 3), our agentic pipeline (model 4), and the agentic pipeline with the refinement step ablated (model 5). 20 participants are then asked to rate the generated answers based on three measures (correctness, time accuracy, and detail). Similarly, we used CLMMs to analyze data as they model the non-interval nature of ratings on a 5-point Likert scale and account for repeated measures across participants and media items through random intercepts.

Descriptive statistics are reported in \Cref{tab:qa_benchmark}. Significant differences were found in all three \textsc{Measures} across different \textsc{Models}. For \textsc{Correctness}, \textsc{Model4} (agentic pipeline with refinement) received the highest ratings, followed by \textsc{Model5}, with all other baselines scoring significantly lower ($p<.001$ for all pairwise comparisons across the five models). For \textsc{TimeAccuracy}, \textsc{Model4} performed best and \textsc{Model5} performed similar to \textsc{Model3}, followed by \textsc{Model2}, and then \textsc{Model1}. Significant differences ($p<.001$) were found in all pairwise comparisons of the five models except between \textsc{Model5} and \textsc{Model3} ($p=.972$).

For \textsc{Detail}, the effect of the refinement step was particularly strong. \textsc{Model4} achieved significantly higher ratings than all other models, followed by \textsc{Model3}. \textsc{Model5} dropped behind \textsc{Model3} and achieved similar ratings as \textsc{Model2}, and \textsc{Model1} came last. Ratings were significantly different in all pairwise comparisons except between \textsc{Model5} and \textsc{Model2}.

These results suggest how the agentic pipeline can improve correctness, temporal accuracy, and detail in the generated structured output of movie files compared with the three baseline models, and how the refinement step is especially crucial to support rich details, and also helps contribute to correctness and temporal accuracy.

\begin{table}[H]
\centering
\caption{Human-scored question answering across 39 questions over 8 films with mean (SD) score values reported for all 20 participants. All models used only the generated textual index as context. Highest mean values are shown in \textbf{bold}.}
\label{tab:qa_benchmark}
\begin{tabular}{lccc}
\toprule
Model & Factual Correctness & Timestamp Accuracy & Detail Completeness \\
\midrule
Gemini 2.5 Pro & 4.36 (1.01) & 4.06 (1.26) & 3.82 (1.16) \\
Gemini 2.5 Flash & 2.82 (1.60) & 2.82 (1.58) & 2.33 (1.26) \\
Gemini 2.0 Flash & 1.51 (0.82) & 1.51 (0.90) & 1.27 (0.52) \\
Ours (Full Pipeline) & \textbf{4.77 (0.49)} & \textbf{4.74 (0.57)} & \textbf{4.46 (0.89)} \\
Ours (w/o Refinement) & 4.01 (1.26) & 4.03 (1.18) & 2.39 (1.36) \\
\bottomrule
\end{tabular}
\end{table}

\subsection{Study 3: Quality of Edited Videos}

\paragraph{Goal.}
We compared our system’s AI-edited movie recaps with human-edited recaps to evaluate genre alignment, editing professionalism, and watchability. This study also examined whether outputs exhibited a recognizable ``AI look.'' The recap genre was chosen because it is both highly popular on YouTube and provides a strong benchmark for narrative comprehension. Recaps require editors to distill long-form, story-driven media into concise, coherent summaries that retain narrative integrity, appropriate pacing, and stylistic consistency.

\paragraph{Materials and Apparatus.}
The evaluation used 8 human-edited recaps and 4 AI-edited recaps generated by our system. All videos were anonymized and uploaded in a uniform format. Playback was standardized to 1080p resolution with audio normalized to --16 LUFS, and participants viewed the videos using the same display and headphones.

\paragraph{Participants.}
Eight experts between the ages of 20 and 30 participated. This group consisted of heavy consumers of recap content, familiar with the stylistic conventions of the genre.

\paragraph{Procedure.}
The viewing order of videos was randomized for each participant, and the source of the recaps (AI vs.\ human) was anonymized. Experts were instructed to rate each video for quality and genre alignment. Optional attention checks were included to stabilize ratings.

\paragraph{Criteria.}
All criteria were scored on a 1--5 scale, where 1 is completely unsatisfactory and 5 is excellent.

\begin{itemize}
    \item \textbf{Genre alignment.} Raters assessed whether the look and feel of the edited video matched the expectations of the recap genre, whether the narration tone was appropriate, whether the music choice supported the style, and whether the video contained suitable introductions and conclusions.
    \item \textbf{Editing professionalism.} Raters evaluated whether cuts were placed at appropriate locations without truncating important action or dialogue, whether pacing emphasized salient events, whether the video was free of glitches or technical errors, and whether the video quality and audio mixing were professional and non-distracting.
    \item \textbf{Watchability.} Raters judged whether the recap provided a coherent explanation of the movie plot and whether the overall video was engaging and enjoyable to watch.
\end{itemize}

\paragraph{Measures.}
Experts scored each video on the rubric and also provided free-text notes identifying any signs of ``AI-ness.'' All logs were anonymized. The analysis procedure and results are described in the following section.

\paragraph{Results.}


Ratings of the 12 movie recap videos (8 edited by humans, 4 edited by agentic pipeline) were analyzed across three \textsc{Categories}, including genre alignment, editing professionalism, and watchability. Similar as before, ratings are provided on a 5-point Likert scale by eight participants, and a CLMM analyzed these 288 ratings (8 participants * 12 videos * 3 categories) with random intercepts for user and video item.

Descriptive statistics are reported in \Cref{tab:video_editing_preference}. The model revealed a strong main effect of editor type on ratings on editing professionalism ($\beta=-3.80, OR=.02, p<.001$), which suggests that the recap videos generated by the agentic pipeline were rated substantially lower than human-edited videos in this dimension. While system-generated recaps are consistently rated lower than human-generated recap videos, this gap is smaller for genre alignment ($\beta=2.56, OR=12.9, p<.001$) and watchability ($\beta=3.00, OR=20.1, p<.001$). This indicates that while the agentic pipeline performed worse on editing quality, participants rated video recaps as comparable to human-generated video recaps in terms of aligning with the recap genre and being watchable. Random effects for user and video item are negligible, which suggests that the findings are consistent across raters and video type.

\begin{table}[H]
\centering
\caption{Mean (SD) values of human ratings (1–5) comparing our AI-edited videos with human-edited recaps. Highest mean values for each category are shown in \textbf{bold}.}
\label{tab:video_editing_preference}
\begin{tabular}{lccc}
\toprule
System & Genre Alignment & Editing Professionalism & Watchability \\
\midrule
Ours (AI-edited videos)   & 4.25 (0.57) & 2.75 (0.80) & 3.94 (0.88) \\
Human-edited (YouTube)    & \textbf{4.66 (0.57)} & \textbf{4.38 (0.97)} & \textbf{4.28 (0.77)} \\
\bottomrule
\end{tabular}
\end{table}

\subsection{Study 4: User Experience Study}

\paragraph{Goal.}
We assessed the usability, intuitiveness, control, and efficiency of our system in comparison to Descript and OpusClip. We also examined prompt reliability, user preferences, and top pain points. These baselines were chosen as they represent the closest currently available AI-enabled editors, although no fully hands-free system exists.

\paragraph{Materials and Apparatus.}
The study used our system’s web UI, with a fallback to a local Temporal server for videos larger than 2GB. Baseline tools included Descript and OpusClip, both accessed via active subscriptions. Participants completed two tasks: (i) create a recap from any pre-downloaded movie, and (ii) create a shortform edit from either the Google I/O 2025 keynote, a three-part online deep learning course, or a set of travel vlog footage. For OpusClip, success was defined as assembling returned clips into a final edited video; if participants were unable to do so, the attempt was marked as a failure.

\paragraph{Participants.}
Fifteen participants aged 17 to 81 were recruited. This convenience sample included individuals with a wide range of technical and video-editing experience, representing diverse professions.

\paragraph{Procedure.}
Each participant used all three tools in randomized order. A brief tutorial and walkthrough were provided for each tool. For our system, participants were explicitly told that it required only one carefully designed prompt, and an example prompt was shown. In contrast, Descript and OpusClip were introduced as timeline-based editors with AI-assisted features. Participants were allowed multiple attempts at each task, with a soft time limit of approximately one hour. All sessions were conducted in person, allowing researchers both to observe participant behavior (such as hesitation points or hidden workarounds) and to avoid complications of remote setup or account management. After completing each tool, participants viewed the output together with the researcher and then filled in a questionnaire.

\paragraph{Measures.}
The questionnaire combined multiple formats: boolean fields recorded whether tasks were completed successfully, partially, or not at all; structured fields captured friction points and criticisms for each tool; open-ended fields elicited participants’ initial impressions and reflections on the usefulness and reliability of natural language prompts (in our system); and ranking fields asked participants to identify which tool was easiest to use and which produced the best final output. Observational notes were also recorded by researchers during each session.


\paragraph{Results.}


The effectiveness of our proposed agentic pipeline was evaluated against two baseline video editing tools (Descript and OpusClip) across Task A (Long-form video recap) and Task B (Short-form video highlight). 15 participants attempted the tasks with all three tools in a within-subjects design, and task success was measured based on whether users achieved the intended edit, with task completion results coded as \textit{Yes}, \textit{Partially}, or \textit{No}.

Friedman tests revealed the following results on task completion. For Task A, there was a significant effect of \textsc{Tool} on completion status ($\chi^2(2) = 20.5$, $p < .001$). Post-hoc Wilcoxon signed-rank tests with Bonferroni correction revealed that our proposed agentic pipeline performed significantly better than both Descript ($p<.01$) and OpusClip ($p<.01$), while the performance of Descript and OpusClip did not differ significantly ($p = .72$). For Task B, Friedman tests also revealed a significant effect ($\chi^2(2) = 9.8$, $p <.01$). Post-hoc Wilcoxon pairwise comparisons suggested that the agentic pipeline significantly outperformed OpusClip ($p = .024$), but the differences between the agentic pipeline and Descript ($p = .169$), as well as between Descript and OpusClip ($p = .591$) were not statistically significant. Together, these task completion results show that participants were more successful in achieving their editing goals with our proposed method.

\begin{table}[h!]
\centering
\caption{Participant feedback organized in themes (participant ID and positive/negative/mixed sentiment shown in brackets).}
\label{tab:themes}
\renewcommand{\arraystretch}{1.5}
\resizebox{\textwidth}{!}{
\begin{tabular}{p{2cm} p{1.4cm} p{13cm}}
\hline
\textbf{Theme} & \textbf{Tool} & \textbf{Feedback Summary} \\
\hline
Ease of Use & Agentic Pipeline & Hands-off (P5, +), simple (P4, +), and easy to use (P1, P3, P5, P6, P7, P10, P11, P15, +), does not require editing knowledge (P3, +). Prompts are expressive (P7, +) but can take time to learn (P1, P9, P11, P15, ±) and require deliberation to achieve desired outcome (P1, ±). \\
 & Descript \& OpusClip & Tedious (P14, -), complicated (P15, -), and required (sometimes too much) manual work (P1, P3, P6, P8, P9, P12, -), required reading information in the side panel (P2, ±), not easy to edit clips (P4, -). Descript auto clip selection does not work well for long videos (P6, P11, -), but the text transcript view in Descript was helpful for video editing (P3, P6, +). \\
\hline
Output Quality & Agentic Pipeline & Reliable and useful (P1 to P10, P15, +), impressive (P2, +), supports fine-grained clipping (P10, +), reliable intro and outro in the video (P13, +), and feels like the tool understands the media (P12, +), but there were occasional short cuts that felt strange (P12, -).  \\
 & Descript \& OpusClip & Both tools provided many good templates (P2, P5, P14, +). OpusClip was good for getting individual short clips but not very useful for editing a whole video (P3, P4, ±). Clip selection was unsuitable for content and poorly sorted (P7, P10, -), and the video clip voice does not match the subtitle displayed (P10, -). Descript was unable to generate automatic voiceover like the agentic pipeline (P5, -), and the removal of pauses during talking can ruin the pacing (P7, -). Clips also lacked logical connections and felt like random selections (P10, -). \\
\hline
User Frustrations & Agentic Pipeline & Slow (P2, P3, P11, -), and lacked templates (P12, -) and display on how much longer the edit would take (P3, P9, -), lacked examples of prompts (P3, P11, -), post-editing options (P10, -), and can't adjust video before it is generated (P5, -), does not allow preview of edited video (P6, -). Pacing of recap can be wrong (P6, P7, P14, -) and multiple tries are often required (P7, P8, -). Sometimes the audio of people talking got cut off (P7, -).\\
 & Descript \& OpusClip & Took a long time to get used to the website (P4, P12, -), vague prompts do not work well (P6, -). Selecting clips and trimming also took a long time (P7, -), felt complicated (P15, -) and very tedious (P14, -). Descript timeline was confusing (P9, -), and the actual editing was still done manually by the user (P12, -). Using OpusClip felt like a job without fun (P12, -). \\
\hline
\end{tabular}
}
\end{table}

A thematic analysis~\cite{guest2012introduction} was applied to identify recurring patterns in the open-ended feedback provided by participants regarding their experiences with each tool, which we summarize in Table~\ref{tab:themes}. These include themes in System Ease of Use, Output Quality, and User Frustrations.
Each theme is
grounded in participant feedback. Participants highlighted the ``hands-off experience'' (P5) and the ``ease of use'' (P3, P7, P10, P11, P15) of our proposed agentic pipeline but noted the extensive manual effort of Descript (P1, P3, P5, P12) and OpusClip (P6, P9, P12). Common user frustrations resulted from slow processing times of our agentic pipeline (P2, P3, P11), difficulty in selecting individual short clips to form the final edited video with Descript (P1, P3, P6, P7, P11) and OpusClip (P6, P7), and the lack of fine-grained control on video edits for the agentic pipeline (P6) and other tools as well (P1). The feedback revealed tradeoffs in the design of AI video editing systems between automation and control/trust. For example, while P11 found the agentic pipeline to ``seemingly think for itself'', P13 suggested how the pipeline lacked mechanisms to allow the user to control a single part of the narration without re-running the entire pipeline. Meanwhile, participants found the two baseline techniques to provide many individual short clips for instant feedback and the video transcript for fine-grained editing control, but disliked the laborious manual editing involved. P1 also expressed how none of the three tools could perfectly balance AI automation and control, which suggests areas for future work.

\subsection{Data Handling and Validity (Across all Studies)}
To ensure transparency and reproducibility, we recorded model versions, prompts, seeds, and hardware details for all experiments. Prompts and rating rubrics will be archived as supplemental materials. All participant IDs and outputs were anonymized. 

We note several potential threats to validity. Convenience sampling may limit generalizability. Familiarity with specific movies could bias rater judgments. Latency and system performance may influence user satisfaction. Finally, results may drift as underlying model versions evolve over time. To mitigate these risks, we randomized and blinded conditions, provided rater calibration with exemplars, and logged visible wait times and progress cues.


\section{Discussion}

The quantitative results highlight not only performance gains but also provide evidence for why an agentic design is particularly effective in this domain. Study 1 confirmed that our agentic pipeline significantly outperformed Gemini 2.0 Flash and Gemini 2.5 Flash in both quality and usability. Compared with Gemini 2.5 Pro, the pipeline achieved a significantly higher usability rating while attaining comparable quality. The odds of our pipeline receiving a higher quality rating were 26× greater than Gemini 2.0 Flash and 20× greater than Gemini 2.5 Flash, while usability odds ratios approached infinity as baselines were consistently rated near the floor. 

In the question-answering task, ratings from Study 2 confirmed that the refinement stage was critical. The full pipeline with refinement achieved the highest scores for correctness, timestamp accuracy, and detail, significantly outperforming all baselines. Ablation revealed that omitting refinement sharply reduced detail ratings, dropping performance close to Gemini 2.5 Flash. This illustrates how iterative self-correction not only improves robustness but also preserves narrative richness.

For video editing quality in Study 3, CLMM analysis revealed a clear tradeoff. Human-edited recaps outperformed AI-edited ones in professionalism, but the gap was smaller in genre alignment and watchability, where system outputs were rated as comparably engaging.

Finally, task completion analysis in Study 4 showed that participants were significantly more successful in achieving long-form recaps with our pipeline than with Descript or OpusClip, with post-hoc tests confirming superiority over both baselines. For short-form tasks, the pipeline significantly outperformed OpusClip and performed comparably to Descript.

Together, these results suggest that modular, agentic orchestration yields measurable improvements in both comprehension and user outcomes. The benefits are not just in mean ratings but in statistical robustness: consistent advantages across models, large odds ratios in usability, and clear significance in pairwise comparisons. These findings support the claim that separating video comprehension into modular stages confers concrete advantages in reliability, correctness, and narrative fidelity.

Several design principles explain this effect. First, the \textit{separation of concerns} allows each sub-agent to focus on a narrow, well-defined task (e.g., narration planning, clip retrieval, beat alignment). This modularity echoes mixed-initiative design guidance and adjustable autonomy—allocating control between human and system in ways that preserve user agency while leveraging automation \cite{horvitz1999mixed,parasuraman2000levels,amershi2019guidelines}. Second, the pipeline incorporates \textit{redundancy and self-correction}: multiple agents contribute at different stages, so outputs are repeatedly checked and refined, mitigating error cascades common in single-pass LLM editing. Third, the system emphasizes \textit{reusability and persistence}. Unlike baseline systems, it natively produces structured intermediate artifacts—scene-level JSON traces, character graphs, and storyboards—that persist across runs. As \emph{boundary objects}, these interpretable artifacts support coordination between users and agents and enable auditing and reuse across tasks. For HCI research, these findings highlight how persistent, interpretable representations can act as boundary objects between human and machine, supporting co-creative iteration.

Bridging current gaps in the literature, our results show (i) how a persistent, time-grounded narrative index complements diffusion editing pipelines that excel at local transformations but lack global story reasoning \cite{molad2023dreamix,qi2023fatezero}; (ii) how modular orchestration and persistent artifacts compensate for limited long-range grounding in current LVLMs—even those with extended context windows \cite{google2025gemini}; and (iii) how narrative-aware planning can guide retrieval and compression to produce supervision signals for efficient media RAG, aligning with broader evidence that retrieval-augmented methods can deliver scalable, high-quality generation \cite{lewis2020rag}. Practically, this architecture helps maintain character arcs, causal alignment, and cross-scene beats at feature-length scale—capabilities not directly targeted by clip-level instruction-tuned models \cite{zhang2023videollama,tsimpoukelli2023m4}.

Several limitations temper the findings of this study. Our user study relied on a convenience sample that, while diverse in age and technical background, may not reflect the practices of professional editors or filmmakers. Future studies should involve longitudinal deployment in real production settings, where issues of trust, efficiency, and integration with existing workflows may surface differently. While the agentic pipeline significantly improved usability and comprehension quality, participants also reported frustration with processing latency, lack of preview mechanisms, and limited opportunities for mid-process correction. Addressing these issues requires both technical advances—such as distilling the semantic index into lightweight retrieval-augmented models—and interaction design advances, such as progressive previews, adjustable autonomy, and user-configurable checkpoints.

Our evaluation also focused on English-language, narrative-heavy media. Extending the pipeline to multilingual content, live streams, or mixed-media projects (e.g., slides and video, game capture) remains an open direction. Similarly, our current indexing emphasizes plot, character, and affective cues; future work may expand to stylistic elements such as cinematography, rhythm, or genre-specific conventions, which are central to many creative practices. Finally, while modularity and redundancy improved robustness, the orchestration of agents is still largely scripted. More adaptive orchestration strategies—where agents dynamically allocate responsibility, escalate uncertainty, or negotiate conflicting outputs—represent a promising future direction. For HCI researchers, this raises an important design question: how much agency should be delegated to the system, and how should users remain in the loop when agents self-coordinate? Exploring this space could lead to richer, more trustworthy co-creative systems.

\section{Conclusion}

In this paper we have presented a prompt-driven, agentic video editing system that enables users to engage with long-form, narrative-rich media entirely through natural language prompts. By constructing a structured semantic index via autonomous comprehension, our pipeline supports editing workflows that would otherwise require extensive manual effort—enabling tasks such as highlight extraction, character-centric retellings, and stylistically coherent summaries across multi-hour footage.

Our evaluations demonstrate that modular, agentic orchestration confers significant advantages over monolithic baselines. The system achieved higher usability ratings, stronger factual and temporal grounding, and higher task success in editing scenarios when compared with both Gemini 2.5 Pro and AI-enabled baseline editors. These results underscore how separation of concerns, redundancy and self-correction, and persistent intermediate artifacts can improve both reliability and user experience in complex creative tasks. For HCI, this highlights the value of designing AI systems not only for output quality but also for interpretability, transparency, and reproducibility of intermediate reasoning.

At the same time, important challenges remain. Participants noted processing latency, lack of preview mechanisms, and limited fine-grained control as recurring frustrations. Extending our pipeline to multilingual, mixed-media, and professional production contexts also requires further work. Future studies should therefore examine longitudinal use in real-world editing environments, explore richer interaction paradigms (e.g., progressive previews, adjustable autonomy), and investigate adaptive orchestration strategies that allow agents to dynamically negotiate responsibility. Addressing these challenges will be essential to fully realizing agentic video editing as a trustworthy and scalable paradigm.

Looking forward, we see this system as a foundation for interactive, model-assisted media editing workflows that merge the reasoning capabilities of foundation models with the intuitiveness of natural language interfaces. By enabling accessible, compositional control over complex media assets—and by foregrounding interpretability as a central design principle—our approach moves toward AI systems that are not only powerful, but also transparent, adaptable, and collaborative partners in creative work.


\section*{LLM Use Disclosure}
We used generative AI tools (ChatGPT) only for language polishing (e.g., grammar, phrasing, and copy-editing) on author-written text.
All study ideas, system design and implementation details, analyses (including statistical modeling and interpretation), figures/tables,
and claims were created and verified by the authors. All AI-suggested edits were reviewed and, when appropriate, revised by the authors.


\bibliographystyle{ACM-Reference-Format}
\bibliography{ref}

\appendix

\end{document}